\documentclass{article}

\usepackage[letterpaper]{geometry}
\usepackage[square,numbers,comma,sort&compress]{natbib}
\usepackage{spconf, amsmath, epsfig}
\usepackage[caption=false]{subfig}
\usepackage[hyphens]{url}
\usepackage[colorlinks, urlcolor=black]{hyperref}
\usepackage{enumitem}
\usepackage{mathrsfs}
\usepackage{graphicx}
\usepackage{multirow, multicol, tabularx, float}
\usepackage{rotating}

\graphicspath{{figs/}}
\DeclareGraphicsExtensions{.pdf,.jpeg,.png,.eps}

\usepackage{fancyhdr}
\begin{document}\sloppy

\twocolumn[{%
\vspace{30mm}
{ \large
\begin{itemize}[leftmargin=2.5cm, align=parleft, labelsep=2.0cm, itemsep=4ex,]

\item[\textbf{Citation}]{M. A. Aabed, G. Kwon and G. AlRegib, "Power of Tempospatially Unified Spectral Density for Perceptual Video Quality Assessment," 2017 IEEE International Conference on Multimedia and Expo (ICME), Hong Kong, 2017, pp. 1476-1481.}

\item[\textbf{DOI}]{https://doi.org/10.1109/ICME.2017.8019333}

\item[\textbf{Review}]{Date of Publication: July 14 2017}

\item[\textbf{Codes}]{\url{https://ghassanalregib.com/perceptual-video-quality-assessment-via-spatiotemporal-3d-power-spectral-analysis/}}

\item[\textbf{Bib}]  {@INPROCEEDINGS\{Aabed2017\_ICME,\\
author=\{M. A. Aabed and G. Kwon and G. AlRegib\},\\ 
booktitle=\{2017 IEEE International Conference on Multimedia and Expo (ICME)\},\\
title=\{Power of Tempospatially Unified Spectral Density for Perceptual Video Quality Assessment\},\\
year=\{2017\},\\
pages=\{1476-1481\}, \}
}

\item[\textbf{Copyright}]{\textcopyright 2018 IEEE. Personal use of this material is permitted. Permission from IEEE must be obtained for all other uses, in any current or future media, including reprinting/republishing this material for advertising or promotional purposes,
creating new collective works, for resale or redistribution to servers or lists, or reuse of any copyrighted component
of this work in other works. }


\item[\textbf{Contact}]{maabed@gmail.com OR \{gukyeong.kwon,alregib\}@gatech.edu\\
\url{http://ghassanalregib.com/}\\}
\end{itemize}
\thispagestyle{empty}
\newpage
\clearpage
\setcounter{page}{1}}}]

\title{Power of Tempospatially Unified Spectral Density for Perceptual Video Quality Assessment}
%
\name{Mohammed A. Aabed, Gukyeong Kwon, and Ghassan AlRegib}
\address{Center for Signal and Information Processing (CSIP)\\
	School of Electrical and Computer Engineering\\
	Georgia Institute of Technology	Atlanta, Georgia 30332, U.S.A.\\
	\{maabed,gukyeong.kwon,alregib\}@gatech.edu}

\maketitle

\begin{abstract}
We propose a perceptual video quality assessment (PVQA) metric for distorted videos by analyzing the power spectral density (PSD) of a group of pictures. This is an estimation approach that relies on the changes in video dynamic calculated in the frequency domain and are primarily caused by distortion. We obtain a feature map by processing a 3D PSD tensor obtained from a set of distorted frames. This is a full-reference tempospatial approach that considers both temporal and spatial PSD characteristics. This makes it ubiquitously suitable for videos with varying motion patterns and spatial contents. Our technique does not make any assumptions on the coding conditions, streaming conditions or distortion. This approach is also computationally inexpensive which makes it feasible for real-time and practical implementations. We validate our proposed metric by testing it on a variety of distorted sequences from PVQA databases. The results show that our metric estimates the perceptual quality at the sequence level accurately. We report the correlation coefficients with the differential mean opinion scores (DMOS) reported in the databases. The results show high and competitive correlations compared with the state of the art techniques.
\end{abstract}
\begin{keywords}
Perceptual quality, video quality, perception, human visual system, 3D power spectral density
\end{keywords}
\section{Introduction}
\label{sec:intro}
The proliferation of visual media, in general, and video streaming services and applications, in particular, in recent years has increased the need for efficient communication, bandwidth and streaming. Video traffic in 2015 accounted for over 55\% of global IP traffic. By 2020, it is estimated that a growth of 68\% in global mobile connections will occur reaching 11.6 billion mobile connections. Mobile video traffic will account for over 75\% of that total. In fact, it will take an individual five million years to watch the amount of monthly video traffic transmitted through global IP networks~\cite{Cisco2016Zeta}. Furthermore, the development and enhancements of video coding standards have been very active over the past decade. In addition to the release of H.265/MPEG-H Part 2 High Efficiency Video Coding (HEVC) in 2013, several development activities from industrial corporations emerged outside the umbrella of Moving Picture Experts Group (MPEG) and International Telecommunications Union (ITU). The recently formed Alliance for Open Media (AOMedia) includes several major industry leaders whose main purpose is developing a true universal royalty-free video coding standard. The alliance is anticipating the release of its first standard in 2017, AOMedia Video 1 (AV1)~\cite{AOMedia,StreamingMedia_State2016}. This also coincides with the Chinese government and companies' ongoing efforts towards AVS2~\cite{AVS}. Nonetheless, video coding development focuses on developing a standard format for the bitstream and decoder mainly. The process involves describing general coding tools without explicitly defining their design. This flexible standardization procedure leaves room for optimizations and innovation but \emph{comes with no guarantees of perceptual video quality}. Henceforth, the importance of quality of experience (QoE) has been critically emphasized in this domain. 

To establish stable video operations and services while maintaining high quality of experience, \emph{perceptual} video quality assessment (PVQA) becomes an essential research topic in video technology. A survey published in 2015 revealed that one out of five viewers will abandon a poor streaming service immediately. Furthermore, 75\% of the users will tolerate a bad stream for up to four minutes before switching to a more reliable one~\cite{Conviva2015}. The significance of PVQA is not limited to quality control only. Perceptual video quality plays a pivotal role in designing and improving super resolution and video enhancements algorithms. PVQA is also conjointly associated with the evolving understanding of the human visual system (HVS) and visual perception in the computational neuroscience community. The two research domains complement one another filling the gaps in our understanding, processing and development of visual media technology. Hence, this paper addresses this problem and introduces a new framework for video quality assessment.

Several video quality assessment approaches have been proposed over the past decade. PVQA has several challenges including the incorporating visual perception characteristics, feature selection and crafting, distortion detection and tracking, and pooling optimization among others. Tempospatial feature processing has been investigated and proposed in several ways in past works. In~\cite{Seshadrinathan10}, the authors use tempospatial Gabor filters and motion trajectory to evaluate spatial, temporal and tempospatial quality of the videos. The work in~\cite{Aabed15} proposed a hierarchal statistical processing model for video quality monitoring using pixel-level optical flow motion fields. Soundararajan and Bovik~\cite{Soundararajan13} utilize natural video statistics in the wavelet domain and entropic differences to predict video quality. Saad~\emph{et.~al}~\cite{Saad2014} proposes a no-reference video quality measure relying on tempospatial natural statistics and motion models using discrete cosine transform (DCT). 3D shearlet transform is applied to videos to capture directions of curvlinear singularities and anisotropic features in~\cite{Li2016}. The authors in~\cite{TorkamaniAzar15} introduce 3D singular value decomposition as content based transform and measure the quality of video by comparing singular values of original and distorted videos. In~\cite{Mittal2016}, a no-reference video quality models of intrinsic statistical regularities observed in natural videos, which are used to quantify distortions. This work introduces a new perceptual video quality framework using tempospatial power spectral density (PSD) processing. To the best of the authors knowledge, our work is the first to explore and utilize PSD in video quality assessment.

In this paper, we propose a new approach to predict the quality of video through the analysis on 3D PSD. We propose a full-reference perceptual video quality metric based on the disruptions in the power of tempospatially unified spectra. Our approach characterizes distortions through the statistical features in 3D PSD by fusing tempospatial power spectral density (TPSD) planes of the distorted and pristine videos to estimate the perceived video quality. The PSD is one of the distinctive frequency domain characteristics of a signal. In addition to the power distribution of video frames, PSD also thoroughly captures scene features and objects~\cite{Torralba03}. Through the 3D processing in our metric, spatial and temporal distortions are analyzed simultaneously. The combined effect from both distortions is effectively captured in the same framework. Moreover, the discrete Fourier transform (DFT) is the main operation required to calculate the PSD, which is a very computationally simple domain transform compared to other operations such as wavelet and curvelet transforms. The computational simplicity enables real-time processing, future extensions and application diversity.

The rest of this paper is organized as follow. We explain statistical features in 3D TPSD  and processing flow of our propose method in Section 2. In Section 3, we validate our metric by examining the correlations with the human mean opinion scores (MOS). We also compare our proposed metric against well-known and state of the art VQA metrics. Section 4 concludes the paper and highlights future directions.

\section{Proposed method}
\subsection{3D Power Spectral Density}
\begin{figure}
	\centering
	{\includegraphics[width=\linewidth]{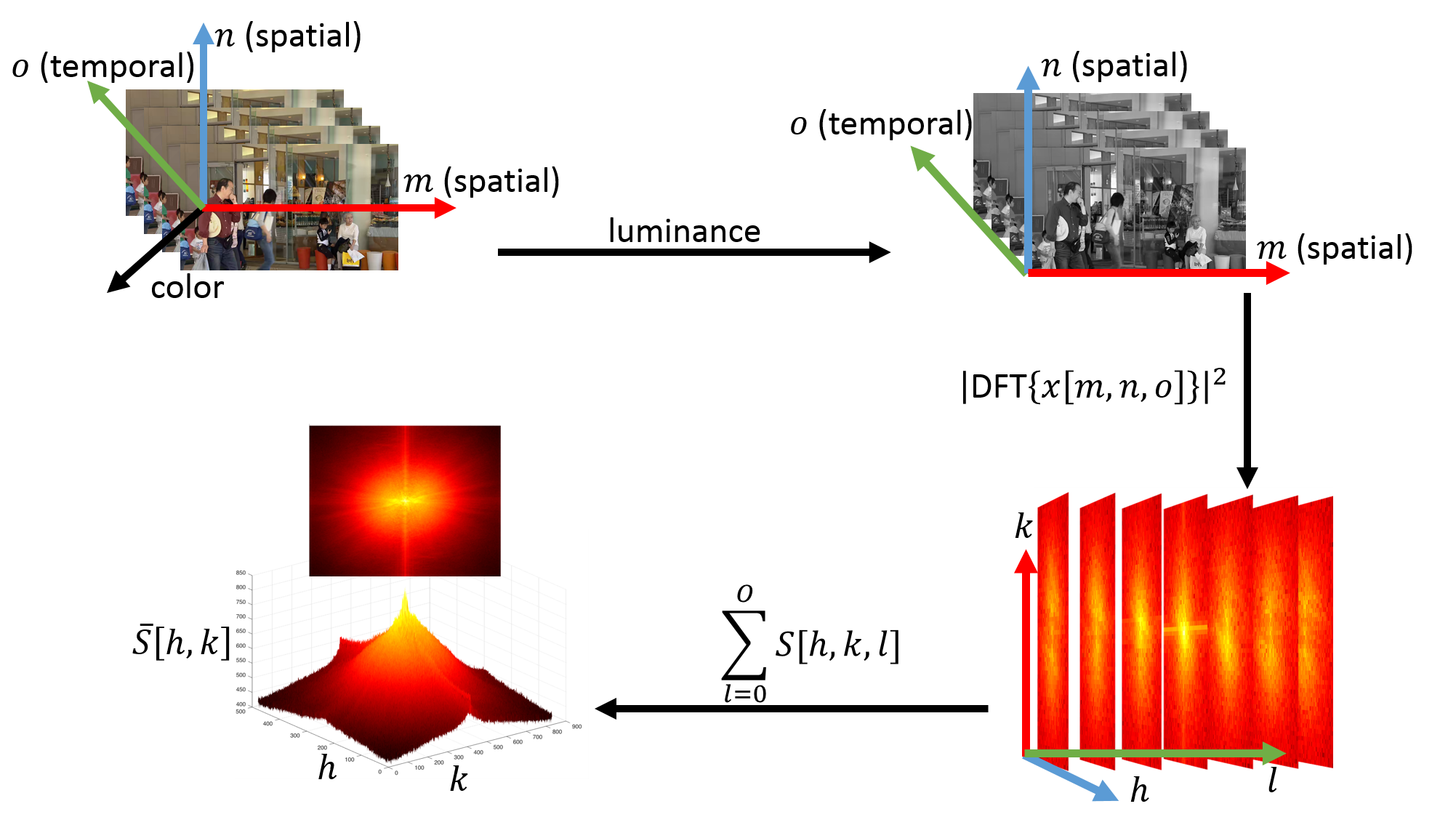}}
	\caption{3D power spectral density tensor-level processing flowchart.\label{fig:PSD_BlockDiagram}}
\end{figure}
A 3D discrete time-space video signal is defined as $x\left[m,n,o\right] \in \mathscr{R}^{M\times N\times O}$, with the one grayscale (luma) channel, where $m$ and $n$ are the spacial indices of the 2D frame and $o$ is the temporal (frame) index. The frequency response of 3D discrete time signal $x\left[m,n,o\right]$ is derived by calculating 3D DFT, $X\left[h, k, l\right] \in \mathscr{C}^{M\times N\times O}$. The 3D discrete PSD, $\mathcal{S}\left[h, k ,l \right] \in \mathscr{R}^{M\times N\times O}$, can be estimated using Parseval's theorem as follows:
\begin{equation}\label{eq:PSD_DFT}
\mathcal{S}\left[h, k ,l \right] = {\frac {1}{M N O}}|X\left[h, k ,l\right]|^{2},
\end{equation}
where $h$, $k$ and $l$ are the discrete frequency indices.

In order to calculate the 2D time-aggregated tempospatial PSD (TPSD) plane at every spatial frequency, $\overline{\mathcal{S}}\left[h, k \right]$, the expression in (\ref{eq:PSD_DFT}) is integrated over the temporal axis, $O$. That is,
\begin{equation}
\overline{\mathcal{S}}\left[h, k \right] = \sum_{l=0}^{O} \mathcal{S}\left[h, k ,l \right],
\end{equation}
where $\overline{\mathcal{S}}\left[h, k \right] \in \mathscr{R}^{M\times N}$. Figure~\ref{fig:PSD_BlockDiagram} illustrates the processing framework for a tensor of frames of size $M\times N \times O$.\vspace{-.3cm}
\begin{figure*}[tbh]
	\centering
	\subfloat[Temporal distortion without motion. Frame 1 in the anchor video is identical to Frame 0.\label{fig:SimplePOTUSSpatial}]{\includegraphics[width=\columnwidth]{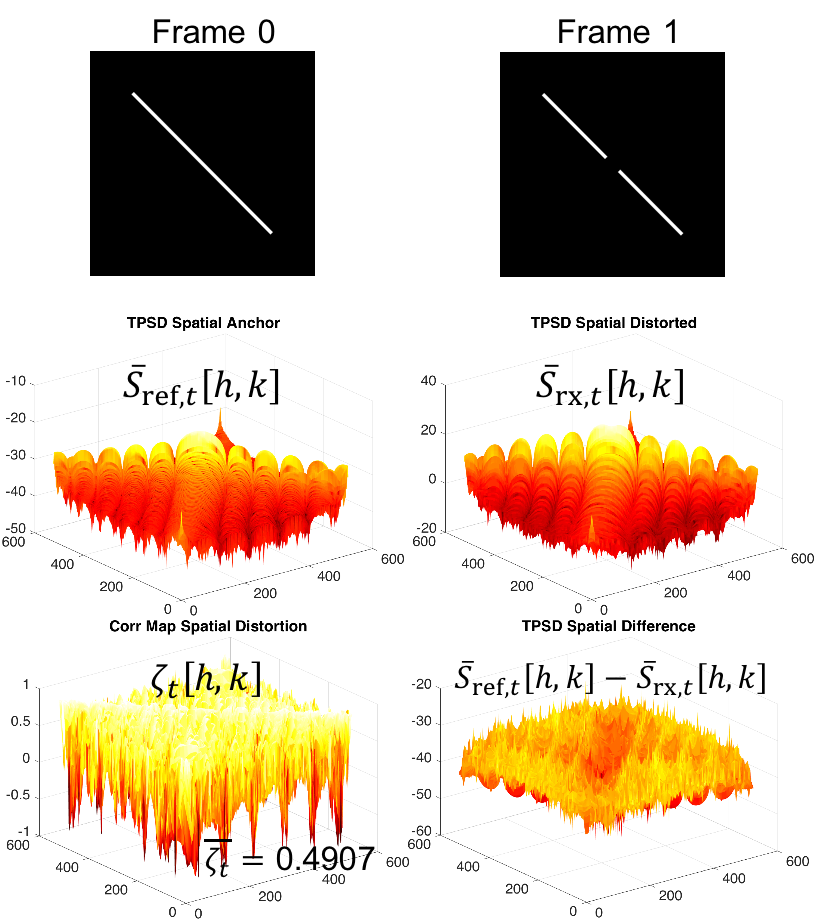}}\hfill
	\subfloat[Temporal distortion with motion. The line in anchor Frame 1 is shifted downwards from its original location in Frame 0 to intorduce a tempospatail change.\label{fig:SimplePOTUSMotion}] {\includegraphics[width=\columnwidth]{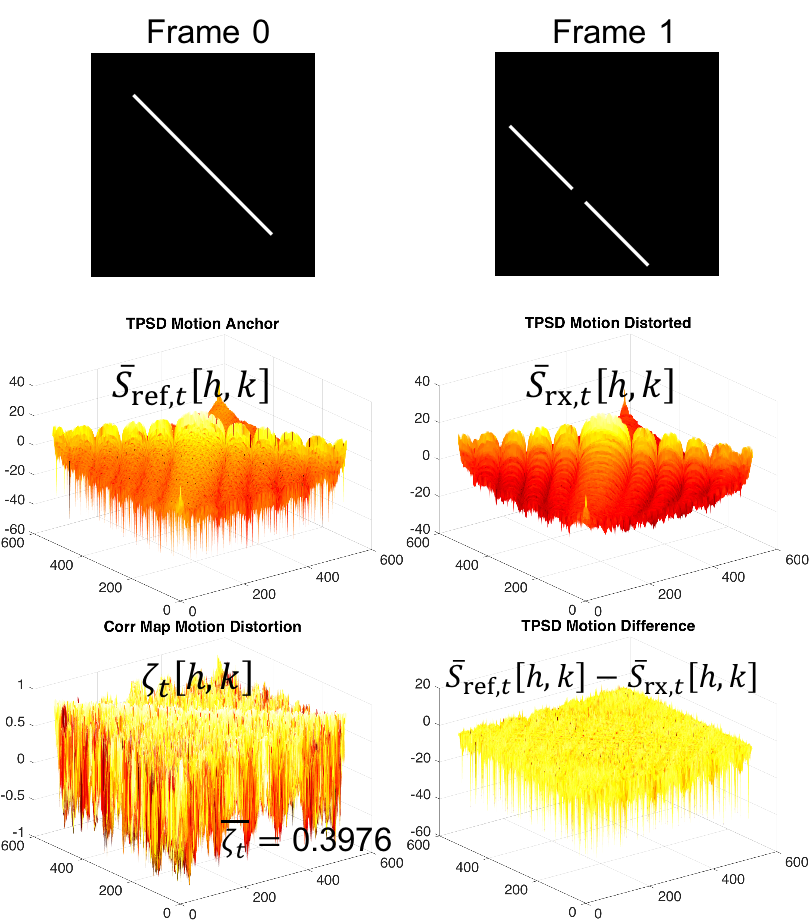}}
	\caption{Simple examples illustrating the principles underlying the proposed metric. Both examples are composed of two frames only ($O=2$) where we show the distorted versions of the Frame 1. For every sequence, we show the the anchor and distorted tempospatial planes, difference map and local-cross correlation distortion map.\label{fig:SimplePOTUS}}
\end{figure*}
\subsection{Video Quality Estimation based on 3D PSD}
Distortions in a video change the tempospatial characteristics in the pixel domain causing a distribution of the signal's PSD. These changes can be captured using 2D time-aggregated TPSD plane, $\overline{\mathcal{S}}\left[h, k \right]$. The deviation of the distorted TPSD from the original free of distortion can be captured in several ways to reflect the change in the energy field. We estimate this variability and map it to perception by measuring a local cross-correlations map between the distorted and anchor TPSD.

Two videos, an anchor video free of distortion and a distorted video, are divided into a set of tensors. For simplicity, we assume the tensors to be of equal size $M\times N\times O$. In practice, tensors sizes may vary depending on coding group of pictures, scene boundaries, processing efficiency, etc. Let the $t^\text{th}$ tensor in the anchor and distorted video be denoted as $x_{\text{ref},t}\left[m,n,o\right]$ and $x_{\text{rx},t}\left[m,n,o\right]$, respectively. Hence, the TPSD planes are denoted by $\overline{\mathcal{S}}_{\text{ref},t}\left[h, k \right]$ and $\overline{\mathcal{S}}_{\text{rx},t}\left[h, k \right]$, respectively.

The local cross-correlations map of the anchor and distorted power spectra, $\zeta_t\left[h,k\right]$, is locally calculated within windows. The local cross-correlations plane is obtained as follows:
\begin{equation}
\zeta_t\left[h, k \right] = \frac{\sigma_{\overline{\mathcal{S}}_{\text{ref},t}\cdot \overline{\mathcal{S}}_{\text{rx},t}}\left[h, k \right]+C}
{\sigma_{\overline{\mathcal{S}}_{\text{ref},t}}\left[h, k \right]\cdot \sigma_{\overline{\mathcal{S}}_{\text{rx},t}}\left[h, k \right]+C} ,
\end{equation}
where
\begin{multline}
\sigma_{\overline{\mathcal{S}}_{\text{X},t}}\left[h, k \right]\\ 
= \sqrt{\sum _{u= -d}^{d}\sum _{v= -d}^{d} \omega_{u, v}({\overline{\mathcal{S}}_{\text{X},t}}\left[h+u, k+v \right]-\mu_{\overline{\mathcal{S}}_{\text{X},t}}\left[h, k \right])^2},
\end{multline}
\begin{multline}
\sigma_{\overline{\mathcal{S}}_{\text{X},t}\cdot \overline{\mathcal{S}}_{\text{Y},t}}\left[h, k \right] \\
= \sum _{u= -d}^{d}\sum _{v= -d}^{d} \omega_{u, v}({\overline{\mathcal{S}}_{\text{X},t}}\left[h+u, k+v \right] - \mu_{\overline{\mathcal{S}}_{\text{X},t}}\left[h, k \right])\\
\times({\overline{\mathcal{S}}_{\text{Y},t}}\left[h+u, k+v \right] - \mu_{\overline{\mathcal{S}}_{\text{Y},t}}\left[h, k \right]),
\end{multline}
and 
\begin{equation}
\mu_{\overline{\mathcal{S}}_{\text{X},t}}\left[h, k \right] =  \sum _{u= -d}^{d}\sum _{v= -d}^{d} \omega_{u, v}{\overline{\mathcal{S}}_{\text{X},t}}\left[h+u, k+v \right].
\end{equation}

${\sigma_{\overline{\mathcal{S}}_{\text{ref},t}\cdot \overline{\mathcal{S}}_{\text{rx},t}}}$ is the cross-covariance, $\mu_{\overline{\mathcal{S}}_{\text{ref},t}}$ and $\mu_{\overline{\mathcal{S}}_{\text{rx},t}}$ are the means,  $\sigma_{\overline{\mathcal{S}}_{\text{ref},t}}$ and $\sigma_{\overline{\mathcal{S}}_{\text{rx},t}}$ are the standard deviations of ${\overline{\mathcal{S}}_{\text{ref},t}}$ and ${\overline{\mathcal{S}}_{\text{rx},t}}$, respectively, and $\omega$ is derived from 2D circular symmetric Gaussian weighting function with the window size of $11\times 11 $ $(d=5)$.

The term $\zeta_t\left[h, k \right]$ in (\ref{eq:PSD_DFT}) defines a 2D tempospatial full-reference perceptual quality map for tensor $t$ of the distorted 
\begin{figure}[t!]
	\centering
	{\includegraphics[width=\linewidth]{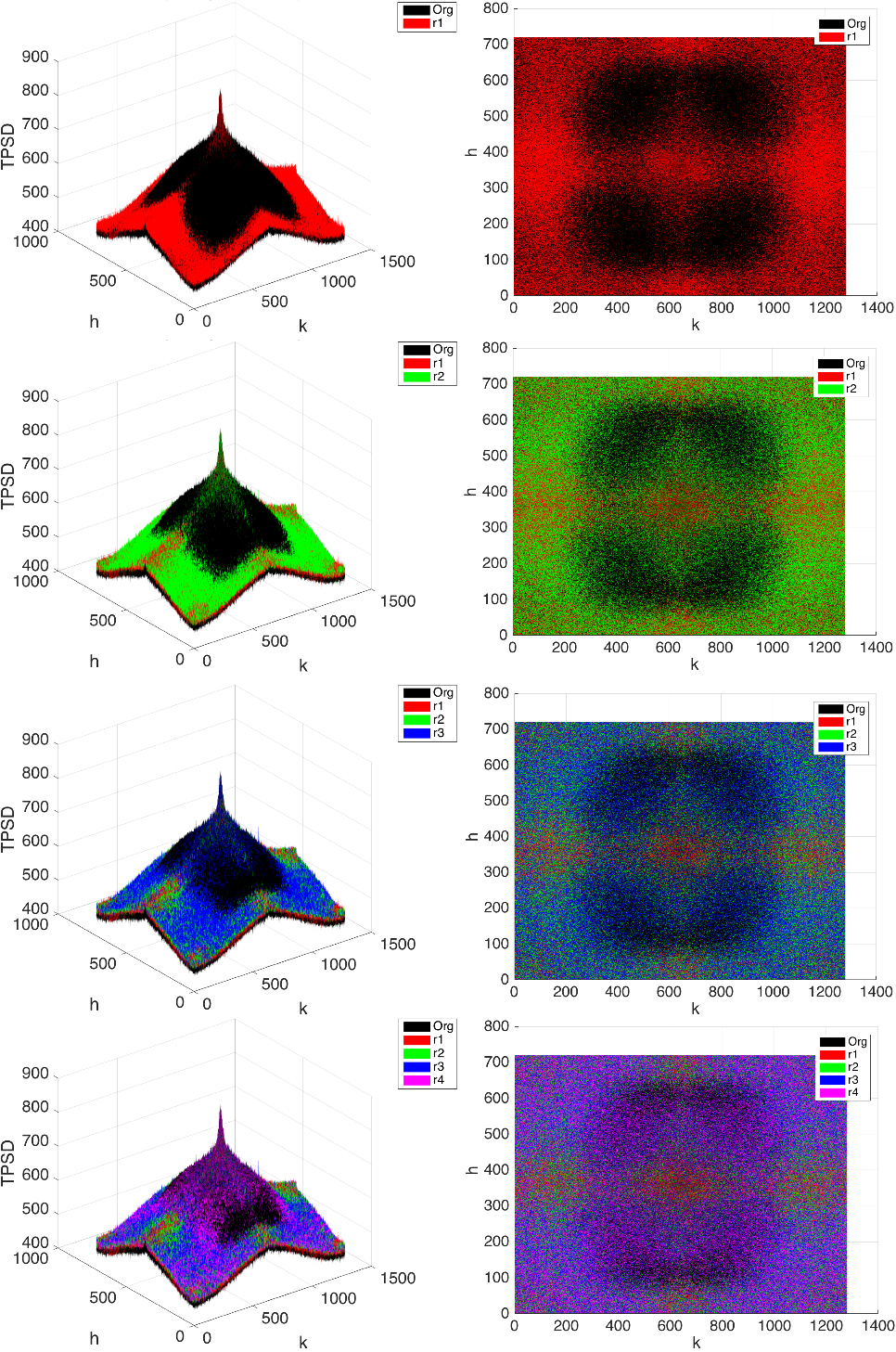}}
	\caption{The incremental change in tempospatial PSD plane for the same video and same set of frames subject to different distortion levels. This example was taken from the Mobile LIVE database, sequence \texttt{Panning Under Oak}, frames $225 - 254$. The distortion magnitudes in the videos are as follows: $r1>r2>r3>r4>\texttt{Org}$ where \texttt{Org} is the anchor video free of distortion.\label{fig:PSD_2DCompare_Incremental}}\vspace{-0.2cm}
\end{figure}
video at every discrete frequency. In our implementation, 30 frames are grouped to form one tensor $(M=1280, N=720, O=30)$ and $C = 4.5 \times 10^{-4}$ is set to prevent instability when denominator is very close zero. The local cross-correlation map, $\zeta_t\left[h, k \right]$, is then averaged to obtain the tensor's perceptual quality score, $\overline{\zeta_t}$, as follows:
\begin{equation}
\overline{\zeta_t} = \frac{1}{MN}\sum _{\forall h}\sum _{\forall k}\zeta_t\left[h, k \right].
\end{equation}

For the temporal pooling of tensors to obtain the overall video quality score, we tested various pooling and statistical processing strategies. Mean pooling was chosen after it has empirically proven its superiority to other functions. Therefore, the overall video quality score is calculated by the average temporal quality of its tensors. That is,
\begin{equation}
\mathcal{P} = \left[\underset{\forall t}{\mathrm{E}}\left[\overline{\zeta_t}\right]\right]^\beta,
\end{equation}
where $\beta$ is an empirically determined sequence-dependent parameter.

Fig.~\ref{fig:SimplePOTUS} shows two simple examples to explain our proposed metric. Both sequences are composed of two frames only. The first frame, Frame 0, is identical in both examples. In Fig.~\ref{fig:SimplePOTUSSpatial}, the second frame, Frame 1, is identical to the first one. Only the distorted version is shown in the figure. In Fig.~\ref{fig:SimplePOTUSMotion}, the edge in Frame 1 is shifted downwards to introduce a simple motion from the previous frame. The distorted version is shown in the figure. We show for each sequence the TPSD, $\overline{\mathcal{S}}_{t}\left[h, k \right]$, for both the distorted and anchor sequences. We also show the difference map between the two as well as local cross-correlations map, $\zeta_t\left[h, k \right]$. Moreover, Figure~\ref{fig:PSD_2DCompare_Incremental} shows the incremental change in the TPSD planes for different levels of distorted tensors with the same contents.

$\zeta_t\left[h, k \right]$ is a local cross-correlations map, which does not evaluate fidelity, it rather examines the contents in a certain frequency and quantifies the cross-correlation or consistency of contents in that frequency neighborhood with original contents. All the temporal and spatial contents corresponding to a certain frequency are unified within this 2D map. Every frequency spectrum in the original contents emits a certain optical energy to stimulate the HVS. A visual distortion will alter this energy in a certain way depending on the nature and severity of the distortion. This in turn causes discomfort and annoyance to viewers. In the context of visual masking, this framework models the visual sensitivity to distortions by estimating the power spectral cross-correlation, where at every frequency this local cross-correlation estimates the human visual discomfort in that frequency neighborhood. Quantifying the cross-correlation of spectral data in every frequency neighborhood measures the masking effect of the original contents (mask) in the presence of distortion (target). In other words, the local cross-correlation acts as a measure of annoyance or discomfort due to disruption of the original power spectrum caused by induced distortion. A high positive correlation indicates the contents to be similar, which yields little to no distortion to the viewer. Low positive and negative local cross-correlation values indicate a degradation in perceptual quality due to distortion. By averaging the map to obtain $\overline{\zeta_t}$, we incorporate the contribution to discomfort from every frequency. This averaging operation penalizes frequency spectra with low positive and negative cross-correlations by reducing the overall average for the whole tensor's perceptual quality.

\section{Experiments and Results}
We validate our proposed metric on the LIVE Mobile Video Quality Assessment database~\cite{Moorthy12}. This database consists of 10 reference videos and 200 distorted videos. All videos are provided in \texttt{YUV420} format. They are 10 seconds in duration at a frame rate of 30 fps and a resolution of 1280 $\times$ 720. 

Every anchor video in the databases has 20 different distorted versions as follows: four different levels of H.264 compression artifacts, four different wireless packet loss levels,  three different rate-adaptation patterns, five different temporal dynamics patterns, and 4 frame-freeze patterns. Compression artifacts were generated by using the JM reference implementation of the H.264 scalable video codec (SVC). Wireless packet loss patterns were simulated by transmitting H.264 SVC compressed video through a simulated wireless channel. Rate adaptation videos have a single rate switch in the video stream. Temporal dynamics videos contain multiple rate switches to test the effect of changes in video quality on the perceived quality. In addition, for the temporal dynamics distortion patterns in the database, we used only the last 210 frames to calculate overall video quality scores. This was motivated by the fact that DMOS values are mostly affected by the last a few seconds of the video~\cite{Richardson2002}. We validated this choice by experimentally verifying that the correlation scores are maximum for all metrics using this range of frames after testing for all other combinations including the full set of frames. 

The performance of our proposed metric is evaluated by calculating correlation scores with the DMOS scores reported in the database. Moreover, we compared our metric with commonly used and state of the art full-reference VQA metrics such as MOVIE~\cite{Seshadrinathan10}, VQM~\cite{Pinson04}, MS-SSIM~\cite{Wang03}, VIF~\cite{Sheikh06}, VSNR~\cite{Chandler07} and NQM~\cite{Damera-Venkata00}. PSNR is included as a baseline VQA metric.  

In addition to the correlation score, we benchmark the computation time to calculate overall video quality score. We chose NQM and VIF to compare the computational time with our proposed method since these two metrics show a comparable performance to the proposed metric in terms of the correlation scores reported in Tables~\ref{table:PCC}-\ref{table:SCC}. All simulations were performed on a Windows PC with Intel Core i7-6700K CPU @ 4.00GHz, 32.0 GB RAM and MATLAB R2015(b).
\begin{figure} [tbh]
	\centering
	{\includegraphics[width=\linewidth]{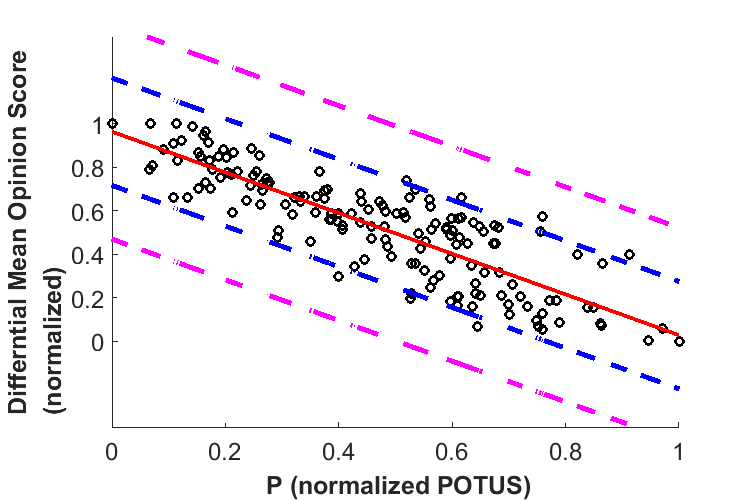}}
	\caption{The predicted quality scores from the metric proposed in this work versus the reported DMOS for the all the sequences in the database. The blue and pink lines are P $\pm \sigma$ and P $\pm 2\sigma$, respectively, where $\sigma$ is the data standard deviation.\label{fig:PSD_scatter_LIVE}}\vspace{-.5cm}
\end{figure}

\subsection{Results and Analysis}
\begin{table}[t]
	\centering
	\caption{Pearson correlation coefficients (PCC) with the DMOS. (Co: Compression, Wl: Wireless channel packet loss, Ra: Rate adaptation, Td: Temporal dynamics)\label{table:PCC}}
	\begin{tabular}{|c|c|c|c|c|c|}
		\hline\hline
		\multirow{2}{*}{Distortion} & \multicolumn{5}{c|}{Pearson Correlation Coefficients} \\ \cline{2-6} 
		        & Co        & Wl        & Ra          & Td        & All    \\ \hline\hline
		PSNR    & 0.784     & 0.762     & 0.536       & 0.417     & 0.691  \\ \hline
		VQM     & 0.782     & 0.791     & 0.591       & 0.407     & 0.702  \\ \hline							    
		MOVIE   & 0.810     & 0.727     & 0.681       & 0.244     & 0.716  \\ \hline                             
		MS-SSIM & 0.766     & 0.771     & 0.709       & 0.407     & 0.708  \\ \hline
		VIF     & 0.883     & 0.898     & 0.664       & 0.105     & 0.787  \\ \hline
		VSNR    & 0.849     & 0.849     & 0.658       & 0.427     & 0.759  \\ \hline
		NQM     & 0.832     & 0.874     & 0.677       & 0.365     & 0.762  \\ \hline
		Proposed & \textbf{0.951}     & \textbf{0.949}     & \textbf{0.856}       & \textbf{0.800}     & \textbf{0.850}  \\ \hline\hline
	\end{tabular}	
\end{table}
\begin{table}[t]
	\centering
	\caption{Spearman correlation coefficients (SCC) with the DMOS. (Co: Compression, Wl: Wireless channel packet loss, Ra: Rate adaptation, Td: Temporal dynamics)\label{table:SCC}}
	\begin{tabular}{|c|c|c|c|c|c|}
		\hline\hline
		\multirow{2}{*}{Distortion} & \multicolumn{5}{c|}{Spearman Correlation Coefficients} \\ \cline{2-6} 
		& Co        & Wl        & Ra          & Td        & All    \\ \hline\hline
		PSNR    & 0.819     & 0.793     & 0.598       & 0.372     & 0.678  \\ \hline
		VQM     & 0.772     & 0.776     & 0.648       & 0.386     & 0.695  \\ \hline							    
		MOVIE   & 0.774     & 0.651     & 0.720       & 0.158     & 0.642  \\ \hline                             
		MS-SSIM & 0.804     & 0.813     & 0.738       & 0.397     & 0.743  \\ \hline
		VIF     & 0.861     & 0.874     & 0.639       & 0.124     & 0.744  \\ \hline
		VSNR    & 0.874     & 0.856     & 0.674       & 0.317     & 0.752  \\ \hline
		NQM     & 0.850     & 0.899     & 0.678       & 0.238     & 0.749  \\ \hline
		Proposed & \textbf{0.959}     & \textbf{0.952}     & \textbf{0.879}       & \textbf{0.811}     & \textbf{0.858}  \\ \hline\hline
	\end{tabular}
\end{table}
\begin{table}[t]
	\centering
	\caption{Computation time to calculate the quality score of 120 frames of \texttt{Harmonicat} video (frames $201-320$).\label{table:Computation time}}
	\begin{tabular}{|c|c|c|c|}
		\hline
		\multirow{3}{*}{Time [sec]} & \multicolumn{3}{c|}{Computation time} \\ \cline{2-4} 
		& VIF        & NQM        & Proposed          \\  \cline{2-4}
		& 255.729     & 59.490     & 15.030      \\ \hline	
	\end{tabular}	
\end{table}
Figure~\ref{fig:PSD_scatter_LIVE} shows the scatter plot of predicted overall video quality score from the proposed method versus DMOS reported in the database. Most of the scatter points are located within one standard deviation boundaries (blue lines). Outliers are also very close to the $\mathcal{P}\pm\sigma$ lines. This shows that our predicted quality scores are highly correlated with the subjective human scores of video quality.

Table~\ref{table:PCC}-\ref{table:SCC} show Pearson's Correlation Coefficients (PCC) and Spearman's Correlation Coefficients (SCC) calculated between predicted quality scores and DMOS in the database. We report PCC and SCC values for each distortion type and with all the videos in the database. The bolded values represent the highest value in each column. For both PCC and SCC, our proposed method outperforms all other VQA metrics by a significant margin for the whole database as well as for every distortion type. In particular, our metric outperforms the second best in compression artifacts (VSNR) by 0.085 in terms of SCC. For wireless packet loss distortions, the proposed metric outperforms the second best (NQM) by 0.053 in terms of SCC. Furthermore, PCC and SCC values from other VQA metrics significantly drop when algorithms are applied to rate adaptation and temporal dynamics distortions. However, our proposed metric shows a robust performance on both distortion types in terms of PCC and SCC. It shows above 0.8 values of PCC, SCC on rate adaptation and temporal dynamics. Since this framework includes both spatial and temporal features via 3D PSD processing, the algorithm effectively captures the impact of dynamic rate changes to human perceived video quality.

Table~\ref{table:Computation time} shows the computational efficiency of our approach compared to other full-reference VQA metrics. This metric only needs 5.88\% of computational time required by VIF and 25.26\% of computational time required by NQM. DFT is one of the simplest domain transform operations and our framework processes a 2D time-aggregated PSD plane for a tensor of frames instead of individual frame processing, which decreases computational burden.

\section{Conclusion}
We propose a full-reference PVQA metric through 3D PSD analysis. In particular, we utilize  2D time-aggregated PSD plane to obtain tempospatial power features and calculate cross-correlation with the reference to quantify the effect of distortion on perceived video quality. We thoroughly evaluate the performance of the proposed metric in terms of correlation with human mean opinion scores of video quality. The results show competitive correlations compared with the state of the art techniques. This work does not make any assumption on coding conditions or video sequence. Furthermore, Our proposed metric has a low computational complexity, which makes it feasible for real-time application. We believe that this work to be a significant step towards understanding the relationship between PSD and perceived quality.

\footnotesize 
\bibliographystyle{IEEEbib}
\bibliography{ICME2017_Mohammed_arXiv}

\end{document}